\documentclass[runningheads]{llncs}
\usepackage{amsmath}
\usepackage{amssymb}
\usepackage[T1]{fontenc}
\usepackage{graphicx}
\usepackage{subcaption}

\usepackage[hidelinks]{hyperref}

\let\oldorcidID\orcidID
\renewcommand{\orcidID}[1]{%
  \href{https://orcid.org/#1}{\oldorcidID{#1}}%
}

\usepackage{color}

\begin{document}

\title{RARF: Region-Aware Rectified Flows for 3D Brain MRI Inpainting}
\titlerunning{RARF for 3D Brain MRI Inpainting}

\author{
Tomas Guija-Valiente\inst{1}\orcidID{0009-0000-0911-3317} \and
Blanca Rodriguez-Gonzalez\inst{1}\orcidID{0009-0007-0982-1293} \and
Norberto Malpica\inst{1}\orcidID{0000-0003-4618-7459} \and
Angel Torrado-Carvajal\inst{1}\orcidID{0000-0002-1540-2809}
}

\authorrunning{T. Guija-Valiente et al.}

\institute{
Medical Image Analysis and Biometry Lab, Universidad Rey Juan Carlos, Madrid, Spain\\
\email{tomas.guija@urjc.es}
}

\maketitle

\begingroup
\renewcommand\thefootnote{}
\footnotetext{Preprint version corresponding to the initial submission prior to peer review. The final accepted version will be openly available in the official MICCAI proceedings on the conference website.}
\addtocounter{footnote}{-1}
\endgroup

\begin{abstract}
Medical image inpainting has the potential to improve automated brain MRI analysis by reconstructing healthy tissue within pathological regions. We introduce RARF, a task-agnostic region-aware rectified flow framework for masked data generation. We instantiate the framework for 3D brain MRI inpainting as our submission to the BraTS Inpainting Challenge 2026. RARF restricts the stochastic interpolation process to the inpainting region, while the observed voxels remain fixed and provide patient-specific anatomical context. A three-dimensional neural network receives the partially voided image, with Gaussian noise filling the missing region, together with the inpainting mask and the corresponding timestep. The model is trained using masked flow-matching and reconstruction-consistency objectives, combined with mask-aware preprocessing and data augmentation. During inference, the learned velocity field transports the initial noise toward a plausible reconstruction of the missing tissue, which is then combined with the unchanged observed anatomy. Experiments under the BraTS evaluation protocol show that the proposed approach produces competitive reconstructions while maintaining anatomical consistency. Source code is available at: \url{https://github.com/TomasGuija/rarf}.

\end{abstract}

\section{Introduction}

Many automated brain Magnetic Resonance Imaging (MRI) analysis methods are designed under the assumption that the input image represents healthy anatomy. This assumption is problematic in neuro-oncology, where healthy images are generally unavailable and downstream tools for tasks such as brain extraction, tissue segmentation, or anatomical parcellation may be unreliable in the presence of lesions~\cite{pollak2025fastsurferlit}. In this context, healthy tissue synthesis can provide a subject-specific anatomical proxy to mitigate pathology-induced bias in subsequent analyses.

The Brain Tumor Segmentation (BraTS) Challenge: Local Synthesis of Healthy Brain Tissue via Inpainting~\cite{kofler2024braintumorsegmentationbrats,baid2021rsnaasnrmiccaibrats2021benchmark} addresses this limitation by formulating healthy tissue synthesis as a standardized inpainting task: given a partially masked brain MRI, the objective is to reconstruct the tumor-affected region with anatomically plausible, tumor-free tissue.

Classical approaches based on diffusion, fast marching, exemplar filling, or smoothness-regularized interpolation have been applied to both natural images and medical imaging problems, including the recovery of missing voxels in brain MRI~\cite{bertalmio2000imageinpainting,telea2004fastmarching,criminisi2004exemplar,torrado}. However, these methods are often insufficient for large pathological regions, where complete anatomical structures must be synthesized rather than locally propagated. Deep learning approaches address this limitation by learning data-driven priors, with mask-aware convolutional and diffusion models proving effective for irregular regions in natural images~\cite{10.1007/978-3-030-01252-6_6,9880056} and for lesion filling, pathology-free reconstruction, and healthy tissue synthesis in medical imaging~\cite{armanious2020ipa,10.1007/978-3-030-59520-3_5,pollak2025fastsurferlit,durrer2024denoising}. In BraTS, existing methods mainly rely on direct U-Net-based reconstruction~\cite{zhang2024synthesis,zhang2025unet,zhang2025robust} or iterative diffusion-based restoration~\cite{durrer2024denoising}. More recently, region-aware diffusion (RAD) has shown the value of adapting the generative process to the mask geometry~\cite{kim2025rad}, which is particularly relevant when synthesizing plausible healthy anatomy while preserving the surrounding patient-specific context.

\begin{figure}[t]
\centering
\includegraphics[width=\linewidth]{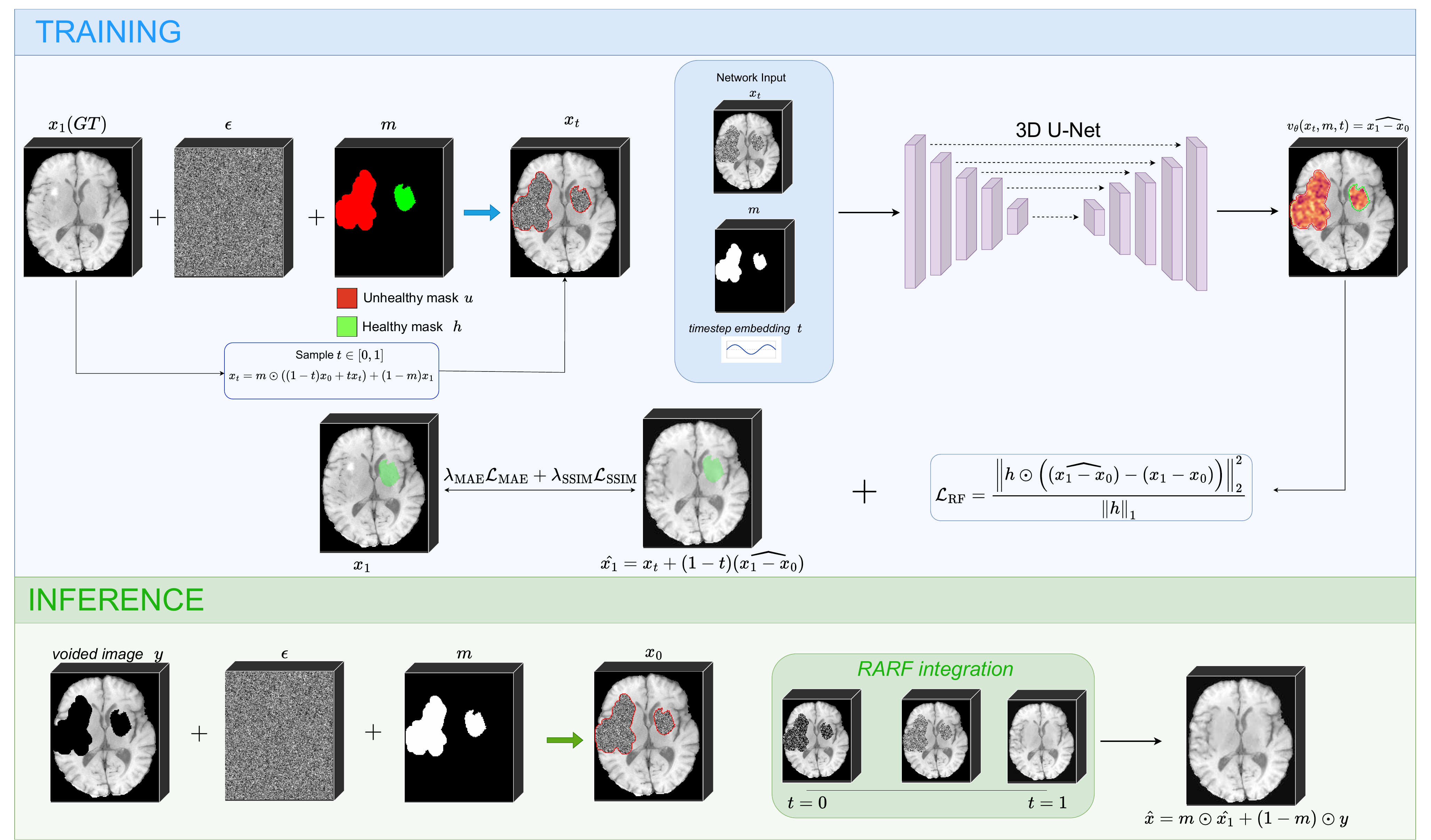}
\caption{
Overview of the BraTS instantiation of our method. During training, RARF interpolation is applied within the inpainting mask while the observed anatomy remains fixed. The network predicts the masked flow velocity from the interpolated image, mask, and timestep, with supervision restricted to healthy tissue. At inference, the missing region is initialized with noise, integrated from $t=0$ to $t=1$, and composed with the unchanged visible anatomy.
}
\label{fig:method_overview}
\end{figure}

Motivated by this region-aware perspective, we propose RARF, a Region-Aware Rectified Flow framework for localized image inpainting~\cite{Liu2022FlowSA}. RARF is designed to be task-agnostic and supports multiple training and inference strategies. In the BraTS instantiation considered here, rectified-flow interpolation and supervision are restricted to the target region, while the observed anatomy is preserved as conditioning context. The model therefore learns to generate plausible healthy tissue within the missing region without modifying the surrounding anatomy. An overview of the resulting training and inference pipeline is shown in Fig.~\ref{fig:method_overview}. Preliminary results suggest that this BraTS-specific instantiation produces anatomically consistent reconstructions.

\section{Methods}

\subsection{Data and pre-processing}
\label{sec: data}
We use the BraTS Local Inpainting dataset~\cite{kofler2024braintumorsegmentationbrats}, derived from the BraTS glioma collection~\cite{baid2021rsnaasnrmiccaibrats2021benchmark}. It contains 1,251 training and 219 validation skull-stripped, co-registered T1-weighted MRI volumes with a common shape of $240\times240\times155$ voxels and $1\mathrm{mm}$ isotropic spacing. Ground-truth volumes are not publicly available for the validation set.

For each training case, the dataset provides the original T1 volume, a healthy tissue mask $h$, an unhealthy tissue mask, and their union $m$, defining the complete inpainting region. The input is voided over $m$, while voxel-wise supervision is restricted to $h$, since tumor intensities are not valid targets for healthy-tissue reconstruction. We use all 1,251 official training cases for optimization. To monitor overfitting without withholding training data, we use the challenge validation cases by placing synthetic masks over visible healthy tissue, creating known regions for supervised validation. The masks are generated and placed following the procedure described in Section~\ref{sec:healthy-mask-augmentation}.

All case volumes and masks are cropped identically using only information available at inference. The foreground is defined as the union of visible nonzero voxels in the voided image and the inpainting mask, and the crop is centered on the midpoint of its axis-aligned bounding box. Volumes are cropped or zero-padded to $166\times196\times152$ voxels.

Intensities are normalized independently for each case. We compute the
$0.5$th and $99.5$th percentiles of the complete voided volume before
cropping, including zero-valued background voxels, and constrain the lower
bound to be nonnegative. Intensities are clipped to these bounds and linearly
mapped to $[0,1]$. During training, the same transformation is applied to the
voided input and clean target.

\subsection{Healthy-mask augmentation}
\label{sec:healthy-mask-augmentation}
Following the mask-generation procedure of Zhang et al.~\cite{zhang2025robust}, we generate five mask variants per training case, including the original healthy mask. Additional masks are created from transformed connected components of training-set lesions, and placed in healthy tissue under constraints on lesion distance, brain coverage, overlap, and diversity. Their union with the unhealthy mask defines the conditioning region, and one variant is sampled per training instance.

\subsection{Region-aware rectified flow}
\label{sec:rarf-formulation}
Let $x_1\in\mathbb{R}^{H\times W\times D}$ denote the normalized
T1 volume, $m\in\{0,1\}^{H\times W\times D}$ the complete inpainting
mask, and $h\in\{0,1\}^{H\times W\times D}$ the healthy-tissue
mask, with $h\subseteq m$. The voided
volume is
\begin{equation}
    y=(1-m)\odot x_1.
\end{equation}

For the BraTS instantiation, inspired by the spatially varying generative process of RAD~\cite{kim2025rad}, we define a localized rectified-flow path in which only the masked region evolves, while the observed context remains fixed during training and inference. Given a Gaussian noise image
\begin{equation}
x_0\sim\mathcal{N}(0,I)
\end{equation}
and a flow time \(t\in[0,1]\), where \(t=0\) denotes the source distribution and \(t=1\) the data distribution, we define
\begin{equation}
x_t
=
m\odot\bigl[(1-t)x_0+t x_1\bigr]
+y.
\label{eq:region_aware_interpolation}
\end{equation}
Thus, interpolation is restricted to \(m\), while the visible anatomy is copied from the voided input.

Equivalently, the interpolation can be expressed through the spatial time map
\begin{equation}
    \tau_t = m\,t+(1-m),
\end{equation}
which assigns time $t$ to masked voxels and fixes visible voxels at the clean
endpoint. Since $\tau_t$ is completely determined by $t$ and $m$, it is not
provided as a separate network input. Instead, the model receives the
channel-wise concatenation
\begin{equation}
    z_t=\operatorname{concat}(x_t,m)
\end{equation}
together with an embedding of the scalar flow time.

The target rectified-flow velocity along the linear path is:
\begin{equation}
\label{eq:target}
    v^\star
    =
    m\odot(x_1-x_0).
\end{equation}

The network is trained to approximate this velocity within the healthy region,
\begin{equation}
    v_\theta(z_t,t)\approx v^\star.
\end{equation}
Although $m$ determines the entire region to be generated, only $h$ contributes to the training objective.

At inference, the region defined by $m$ is initialized with Gaussian noise and the learned velocity field is integrated from $t=0$ to $t=1$. At every integration step, the predicted update is multiplied by $m$, and the observed voxels are restored from $y$. Consequently, the visible anatomical context remains unchanged throughout generation.

\subsection{Network architecture}

We parameterize the velocity field using a 3D U-Net. Its input
has two channels, corresponding to the current state $x_t$ and the complete
mask $m$, and its output is a single-channel velocity volume with the same
spatial dimensions.

The encoder has three resolution levels with channel widths $32$, $64$, and
$128$. Each level contains two residual blocks composed of group normalization,
SiLU activations, and $3\times3\times3$ convolutions. Downsampling is performed
using stride-two $3\times3\times3$ convolutions. The decoder mirrors the encoder
and combines its features with the corresponding encoder activations through
U-Net skip connections. Upsampling uses nearest-neighbor interpolation followed
by a $3\times3\times3$ convolution.

The scalar flow time is multiplied by $T=1000$ and encoded using sinusoidal
features. A two-layer multilayer perceptron maps this encoding to a
$128$-dimensional time representation, which is projected and added to every
residual block. The final prediction head applies group normalization, a SiLU
activation, and a zero-initialized $3\times3\times3$ convolution.

\subsection{Training objective}

Although the complete mask $m$ defines the region to be synthesized,
supervision is restricted to the healthy-tissue mask $h\subseteq m$. This
prevents pathological tissue from contributing directly to the learning
objective. We define the masked mean-squared and mean-absolute errors as

\begin{equation}
    \operatorname{MSE}_{h}(a,b)
    =
    \frac{\left\|h\odot(a-b)\right\|_2^2}{\left\|h\right\|_1},
    \qquad
    \operatorname{MAE}_{h}(a,b)
    =
    \frac{\left\|h\odot(a-b)\right\|_1}{\left\|h\right\|_1}.
\end{equation}

For the noise sample $x_0\sim\mathcal{N}(0,I)$ and clean target $x_1$, the
region-aware interpolation has the constant target velocity $v^\star$ defined
in Eq.~\ref{eq:target}. The primary rectified-flow objective is therefore 

\begin{equation}
    \mathcal{L}_{\mathrm{RF}}
    =
    \operatorname{MSE}_{h}
    \left(
        v_\theta(z_t,t),
        v^\star
    \right).
\end{equation}

We supplement this squared flow-matching objective with voxel-wise and
structural auxiliary terms. Given the predicted velocity, the corresponding
clean endpoint is estimated as
\begin{equation}
    \widehat{x}_1
    =
    x_t+(1-\tau_t)\odot v_\theta(z_t,t).
\end{equation}
Along the linear region-aware path, its endpoint error can be written as
\begin{equation}
    \widehat{x}_1-x_1
    =
    (1-\tau_t)\odot
    \left(v_\theta(z_t,t)-v^\star\right).
\end{equation}
To remove the dependence of the endpoint error on the remaining trajectory length, we define the normalized endpoint error
\begin{equation}
    \widetilde{e}_t
    =
    \frac{\widehat{x}_1-x_1}
         {\max(1-\tau_t,\epsilon)},
    \qquad
    \epsilon=10^{-3},
\end{equation}
where the maximum and division are applied element-wise. Within the
supervised region and away from the numerical clamp,
$\widetilde{e}_t$ is equivalent to
$v_\theta(z_t,t)-v^\star$.

We apply a masked mean-absolute error to this normalized endpoint error:
\begin{equation}
    \mathcal{L}_{\mathrm{MAE}}
    =
    w(t)\operatorname{MAE}_{h}
    \left(
        \widetilde{e}_t,0
    \right),
    \qquad
    w(t)=1+\frac{1}{2}t^2.
\end{equation}
This term complements the squared flow-matching objective with an $\ell_1$
penalty, providing direct voxel-wise supervision while reducing the influence
of isolated large errors.

To additionally encourage preservation of local anatomical structure and
contrast, we compare the estimated and target endpoints using a masked
structural-similarity loss:
\begin{equation}
    \mathcal{L}_{\mathrm{SSIM}}
    =
    w(t)\operatorname{mean}_{h}
    \left[
        1-
        \operatorname{SSIM}_{\mathrm{map}}
        \left(
            \widehat{x}_1,x_1
        \right)
    \right].
\end{equation}
Before computing the structural similarity index measure (SSIM), both endpoints are clipped to the normalized intensity
range $[0,1]$ and set to zero outside $h$. The resulting SSIM map is averaged
only over the supervised region.

The weighting factor $w(t)$ mildly emphasizes later trajectory stages, where
the evolving sample is closer to the clean endpoint and fine reconstruction
errors become more relevant. Training times are sampled from a mixture
distribution: with probability $0.75$, $t$ is sampled uniformly from
$\mathcal{U}(0,1)$; otherwise, it is sampled as
$t=1-u^2$, with $u\sim\mathcal{U}(0,1)$, increasing the frequency of examples
near the clean endpoint.

The complete training objective is
\begin{equation}
    \mathcal{L}
    =
    \mathcal{L}_{\mathrm{RF}}
    +
    \lambda_{\mathrm{MAE}}\mathcal{L}_{\mathrm{MAE}}
    +
    \lambda_{\mathrm{SSIM}}\mathcal{L}_{\mathrm{SSIM}}.
\end{equation}

\subsection{Inference}
\label{sec:inference}

At inference, the masked region is initialized with Gaussian noise and the
learned velocity field is integrated from $t=0$ to $t=1$. After each update, voxels outside the mask are restored from the observed image.

The number of integration steps balances accuracy and runtime. Based on validation results, we use four integration time points with midpoint integration, which provided the best empirical performance while remaining substantially faster than conventional denoising diffusion probabilistic model (DDPM) sampling.

Since different initial noise realizations may produce different plausible
completions, we also consider drawing $K$ independent samples
$\{\hat{x}^{(k)}\}_{k=1}^{K}$. Their voxel-wise average is
\begin{equation}
    \bar{x}
    =
    \frac{1}{K}\sum_{k=1}^{K}\hat{x}^{(k)}.
\end{equation}
This Monte Carlo estimate approaches the conditional mean, which minimizes expected squared error and can therefore improve distortion-based metrics such as MSE and PSNR. However, when multiple anatomical completions are plausible, it may blur fine structures, illustrating the distortion--perception trade-off in image restoration~\cite{NEURIPS2021_d77e6859}.

To retain more sample-level detail, we evaluate sample selection
strategies. First, we select the generated sample closest to the sample mean within the inpainting mask:
\begin{equation}
    \hat{x}_{\mathrm{medoid}}
    =
    \arg\min_{\hat{x}^{(j)}}\,
    \operatorname{MSE}_{m}\left(\hat{x}^{(j)},\bar{x}\right),
\end{equation}
where $\operatorname{MSE}_{m}$ denotes the MSE restricted to the complete
inpainting mask.

We also evaluate kernel density steering (KDS), an inference-time scaling method
for image restoration based on mode seeking in the sample distribution
~\cite{DBLP:conf/nips/HuMSKMD25}. KDS jointly evolves multiple samples and, at each integration step, compares their predicted clean endpoints using a Gaussian kernel. A mean-shift update then steers the samples toward regions of higher estimated density. After integration, we compute the voxel-wise mean
of the final particles within the inpainting region and select the particle
with the smallest mean-squared distance to this mean.

Finally, we consider a minimum Bayes-risk (MBR) selection strategy inspired by agreement-based decoding~\cite{natsumi-etal-2025-agreement}. Each generated completion is compared with the remaining samples using a weighted combination of masked MSE and SSIM, and the candidate with the lowest average pairwise risk is selected. This yields a central representative completion without averaging voxel intensities.

Section~\ref{sec:inference_ablation} compares these inference strategies using
the same trained checkpoint, isolating their effects from differences in model
training.

\section{Results}
\label{sec:results}

\subsection{Experimental setup and metrics}

We conduct all experiments on the BraTS Local Inpainting dataset described in
Sec.~\ref{sec: data}, using the training and local-validation setup,
preprocessing pipeline, and mask-generation strategy introduced therein. Since the official validation targets are withheld, local validation is used to monitor training and select model configurations. Inference ablations are evaluated through separate submissions to the official Synapse platform using the same trained checkpoint. We report mean squared error (MSE), mean absolute error (MAE), peak signal-to-noise ratio (PSNR), and structural similarity index measure (SSIM).

\subsection{Ablation study}
\label{sec:inference_ablation}

Our ablation study focuses on the inference strategies described in Sec.~\ref{sec:inference}, all evaluated using the same trained checkpoint. All reported results use an exponential moving average (EMA) of the model parameters, with decay $\beta=0.999$. We compare them quantitatively through separate submissions to the Synapse evaluation platform and qualitatively through representative reconstructed volumes. Ablations of the broader training modes supported by RARF are left for future work, as they fall outside the scope of this challenge submission.

\begin{table}[h]
\centering
\caption{Inference ablation results on the official BraTS validation set.}
\label{tab:inference_ablation}
\begin{tabular}{lcccc}
\hline
\textbf{Inference strategy}
& \textbf{MSE $\downarrow$}
& \textbf{MAE $\downarrow$}
& \textbf{PSNR $\uparrow$}
& \textbf{SSIM $\uparrow$} \\
\hline
30-sample average         & \textbf{0.006} & \textbf{0.018} & \textbf{23.830} & \textbf{0.823} \\
30-sample closest to mean & 0.009 & 0.021 & 21.911 & 0.780 \\
KDS                       & 0.009 & 0.021 & 22.096 & 0.784 \\
MBR                       & 0.009 & 0.022 & 21.982 & 0.781 \\
Regular inference         & 0.010 & 0.022 & 21.942 & 0.777 \\
\hline
\end{tabular}
\end{table}

Table~\ref{tab:inference_ablation} shows that distortion-based metrics benefit from averaging multiple samples, consistent with interpreting the sample mean as a Monte Carlo estimate of the conditional expectation. However, averaging also produces smoother reconstructions and may blur fine anatomical structures, illustrating the distortion--perception trade-off. Representative-sample strategies provide a compromise by improving distortion metrics over single-sample inference while better preserving sample-level detail. Nevertheless, with \(K=30\), all multi-sample strategies require substantially greater inference time.

\begin{figure*}[!t]
\centering

\captionsetup[subfigure]{
    labelformat=empty,
    font=scriptsize,
    justification=centering,
    singlelinecheck=true,
    skip=2pt
}

\begin{minipage}[t]{0.88\textwidth}
\vspace{0pt}
\centering

\begin{subfigure}[t]{0.158\linewidth}
\centering
\includegraphics[width=\linewidth]{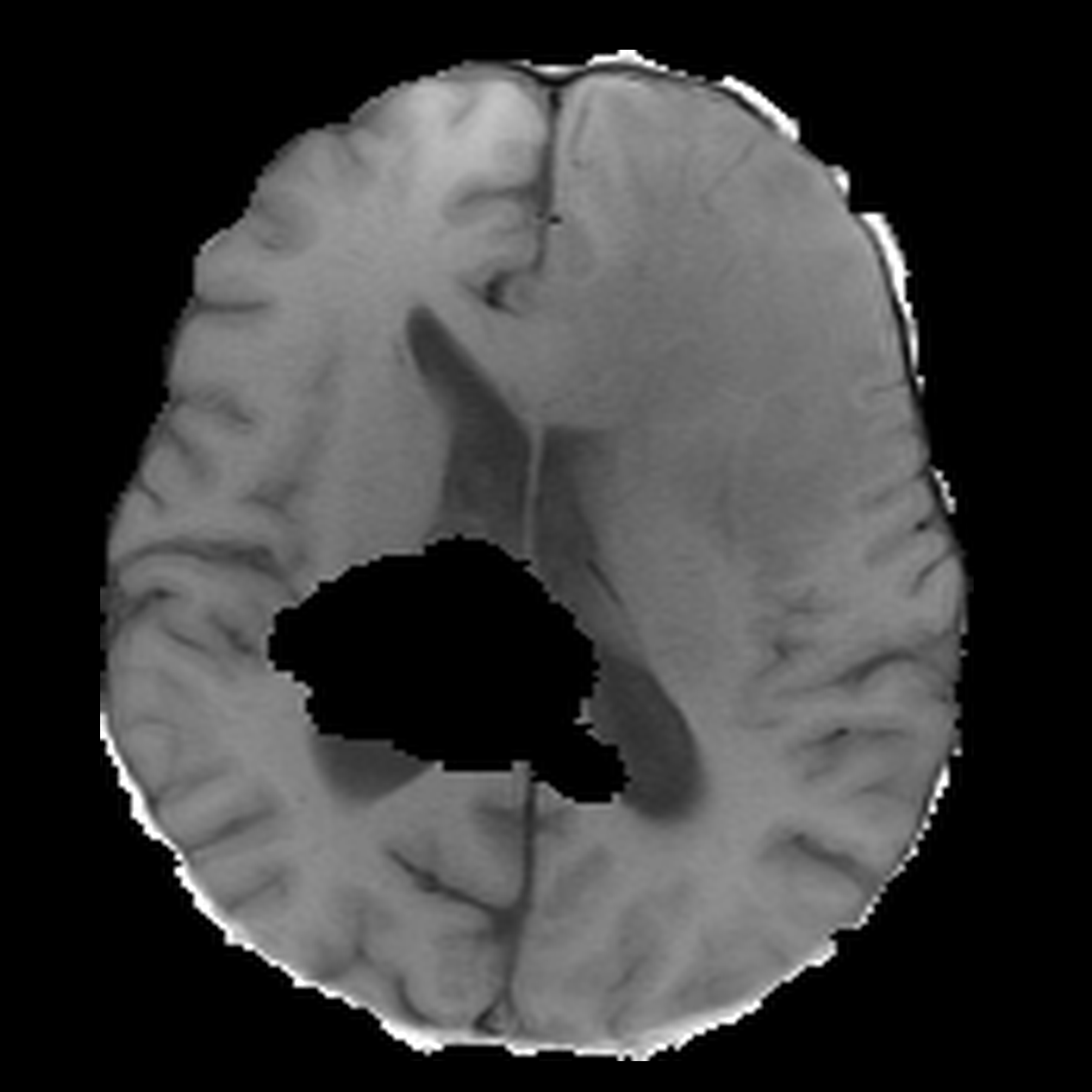}
\end{subfigure}
\hfill
\begin{subfigure}[t]{0.158\linewidth}
\centering
\includegraphics[width=\linewidth]{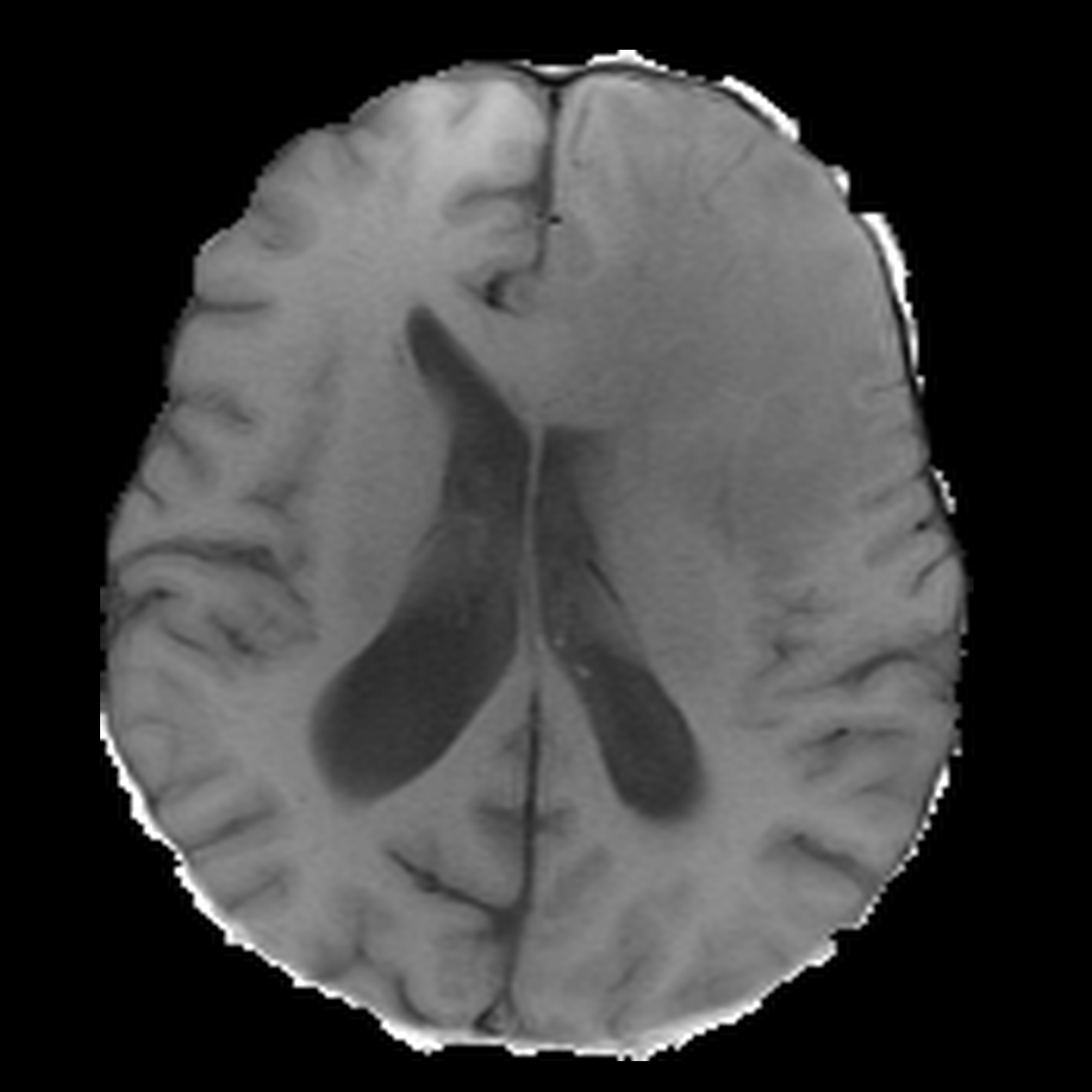}
\end{subfigure}
\hfill
\begin{subfigure}[t]{0.158\linewidth}
\centering
\includegraphics[width=\linewidth]{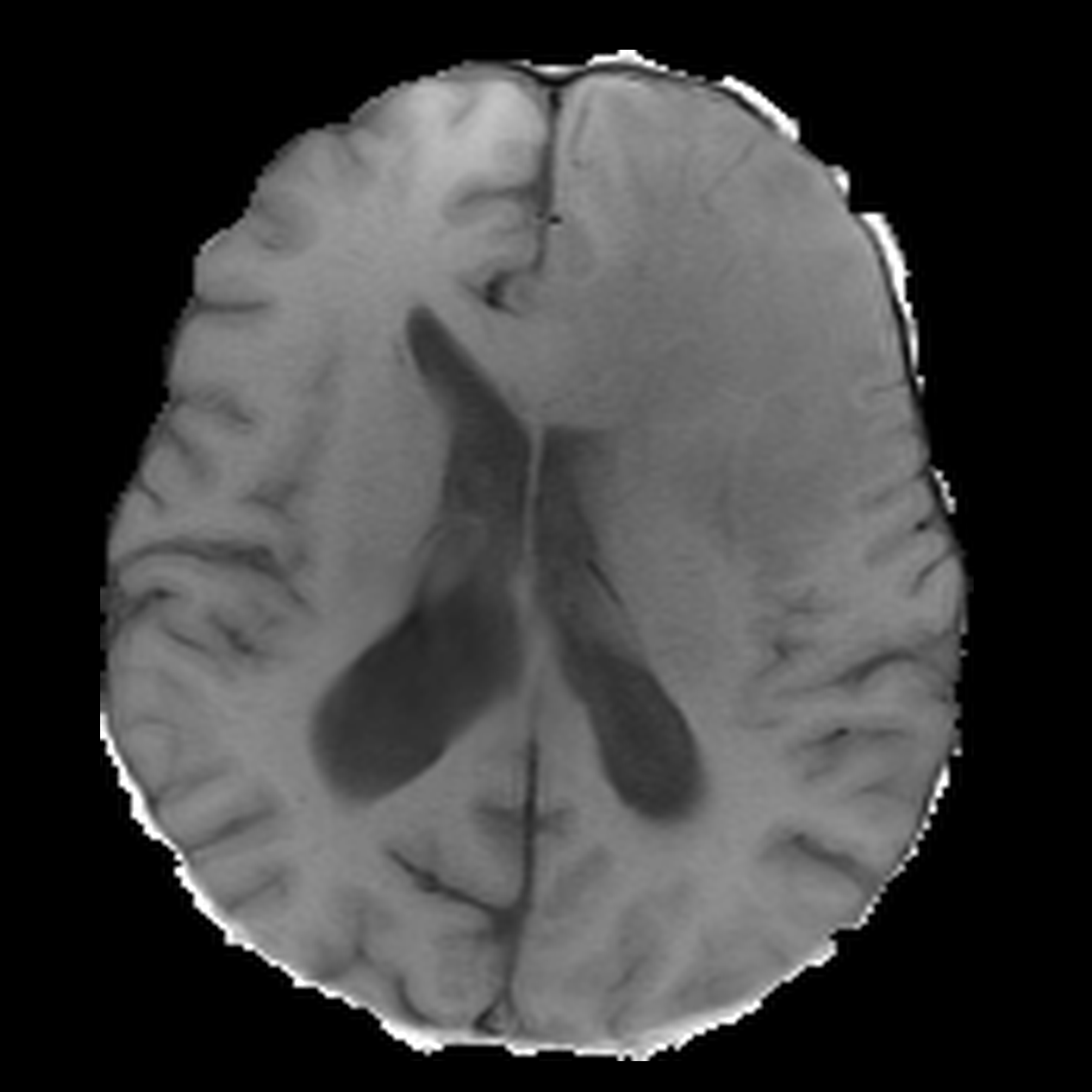}
\end{subfigure}
\hfill
\begin{subfigure}[t]{0.158\linewidth}
\centering
\includegraphics[width=\linewidth]{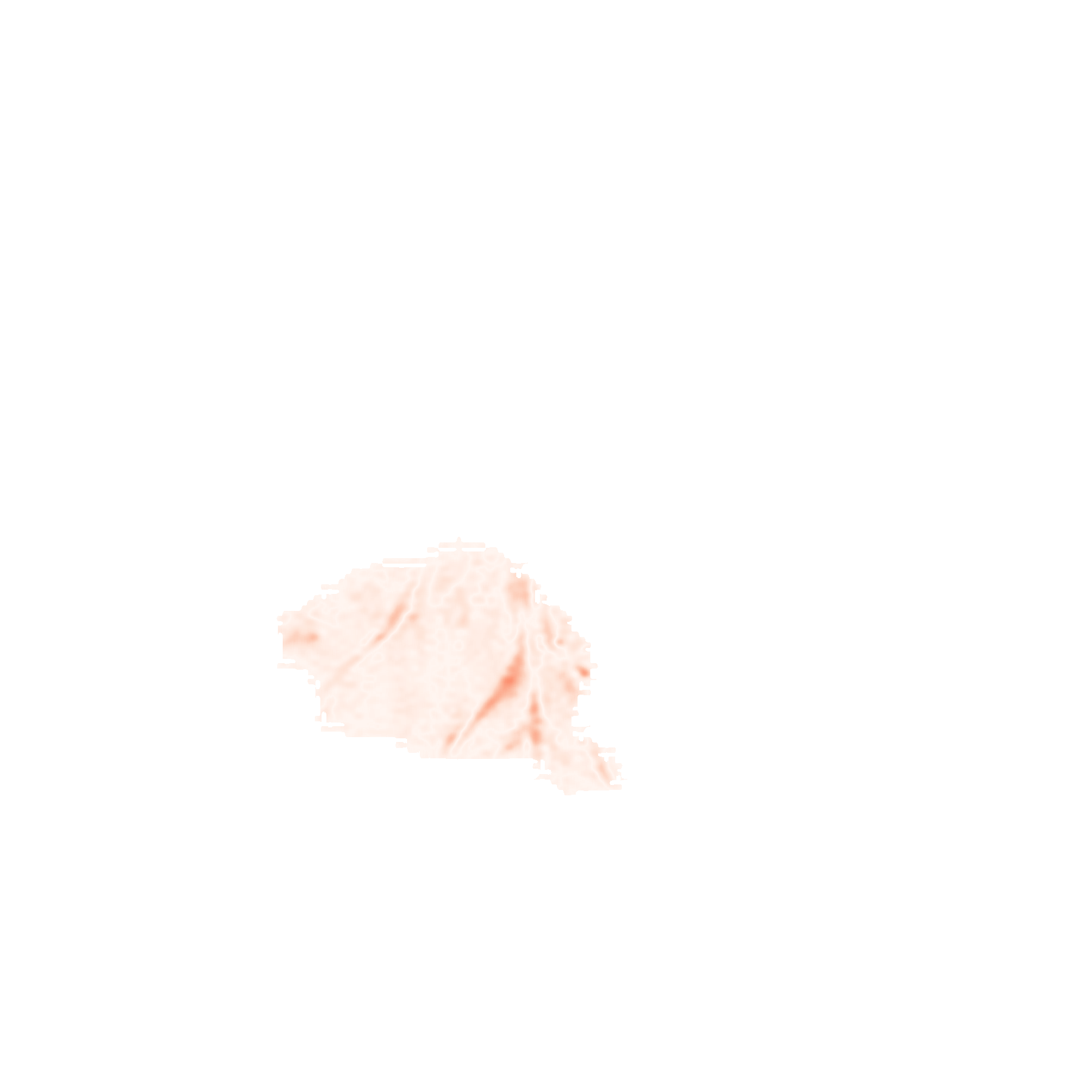}
\end{subfigure}
\hfill
\begin{subfigure}[t]{0.158\linewidth}
\centering
\includegraphics[width=\linewidth]{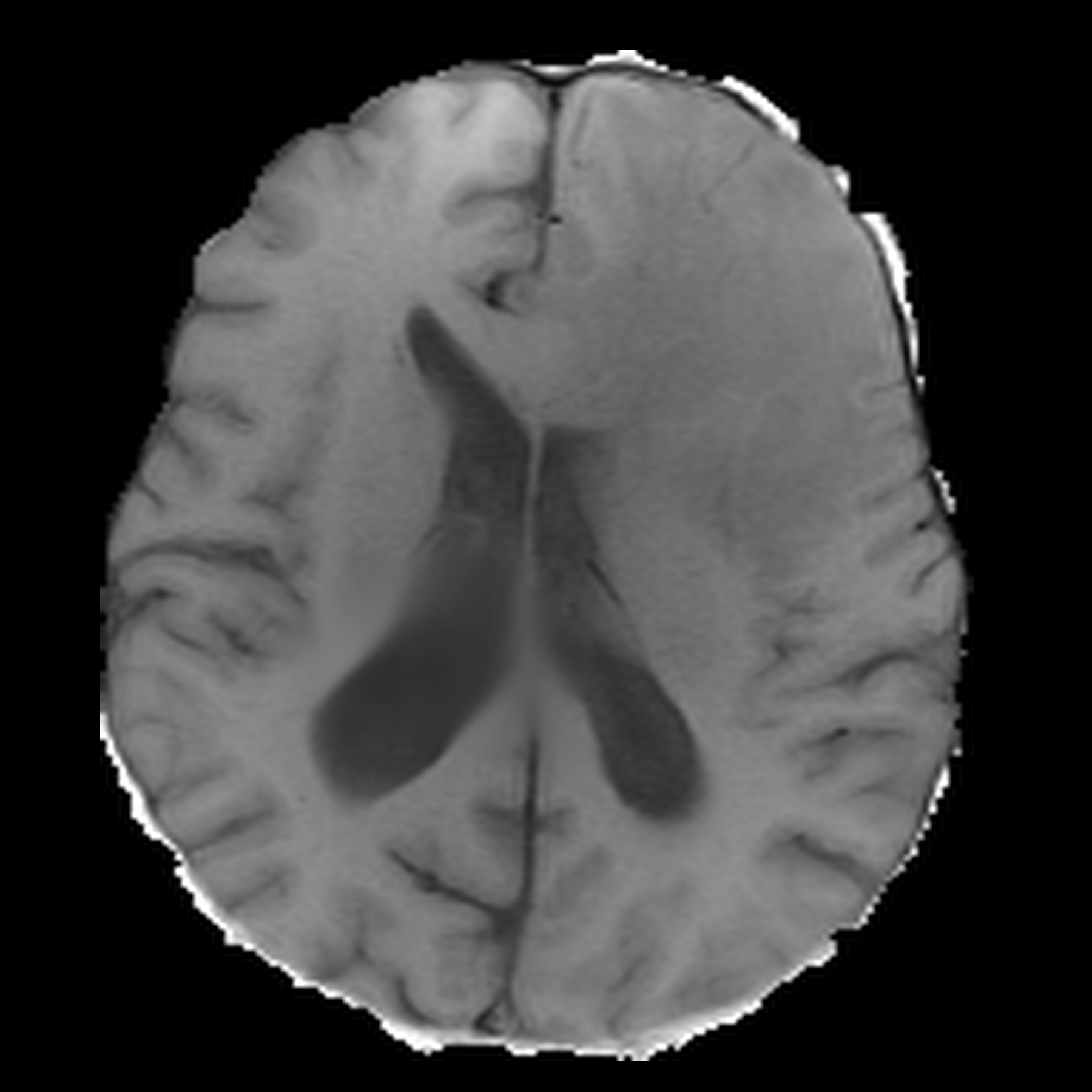}
\end{subfigure}
\hfill
\begin{subfigure}[t]{0.158\linewidth}
\centering
\includegraphics[width=\linewidth]{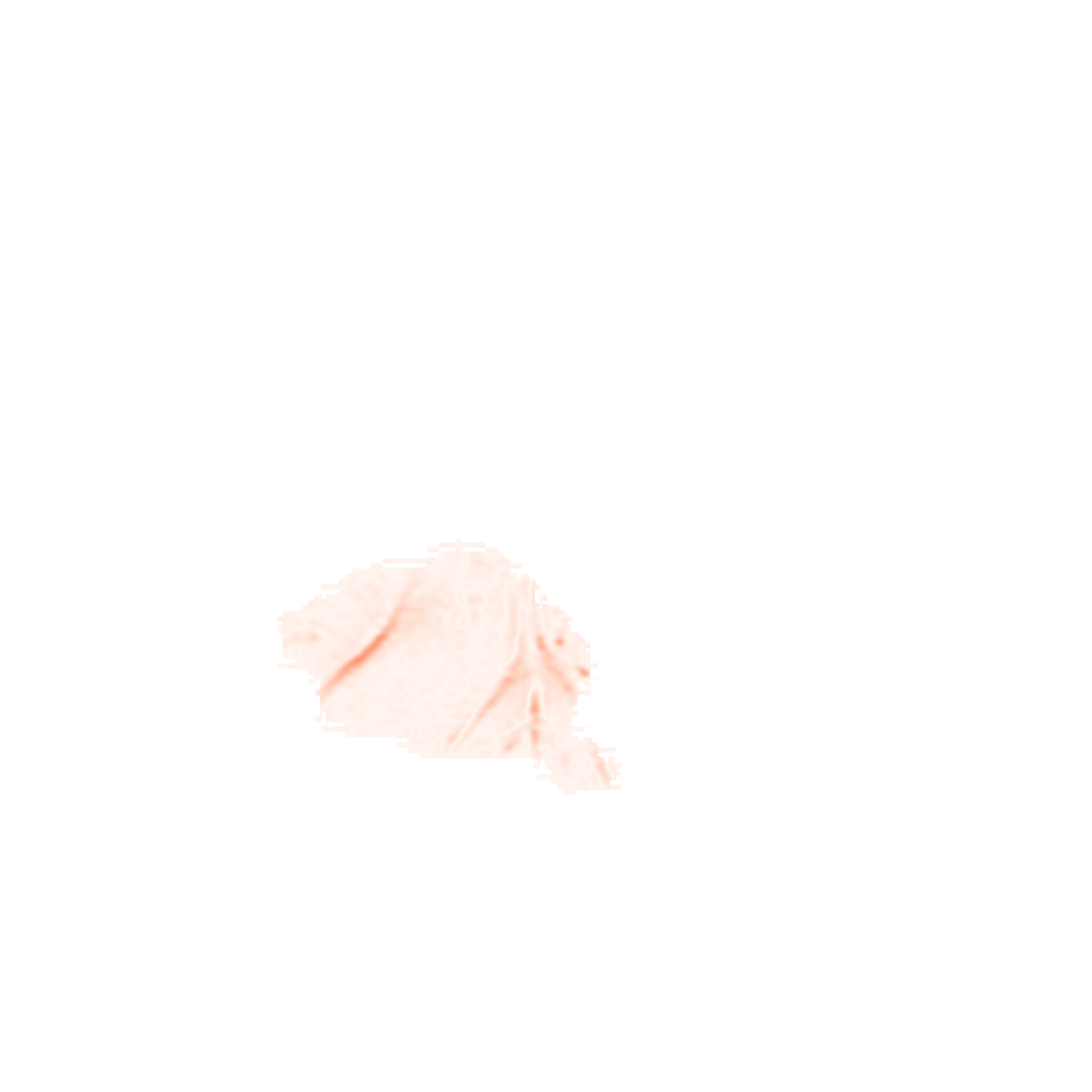}
\end{subfigure}

\par
\vspace{0.6em}

\begin{subfigure}[t]{0.158\linewidth}
\centering
\includegraphics[width=\linewidth]{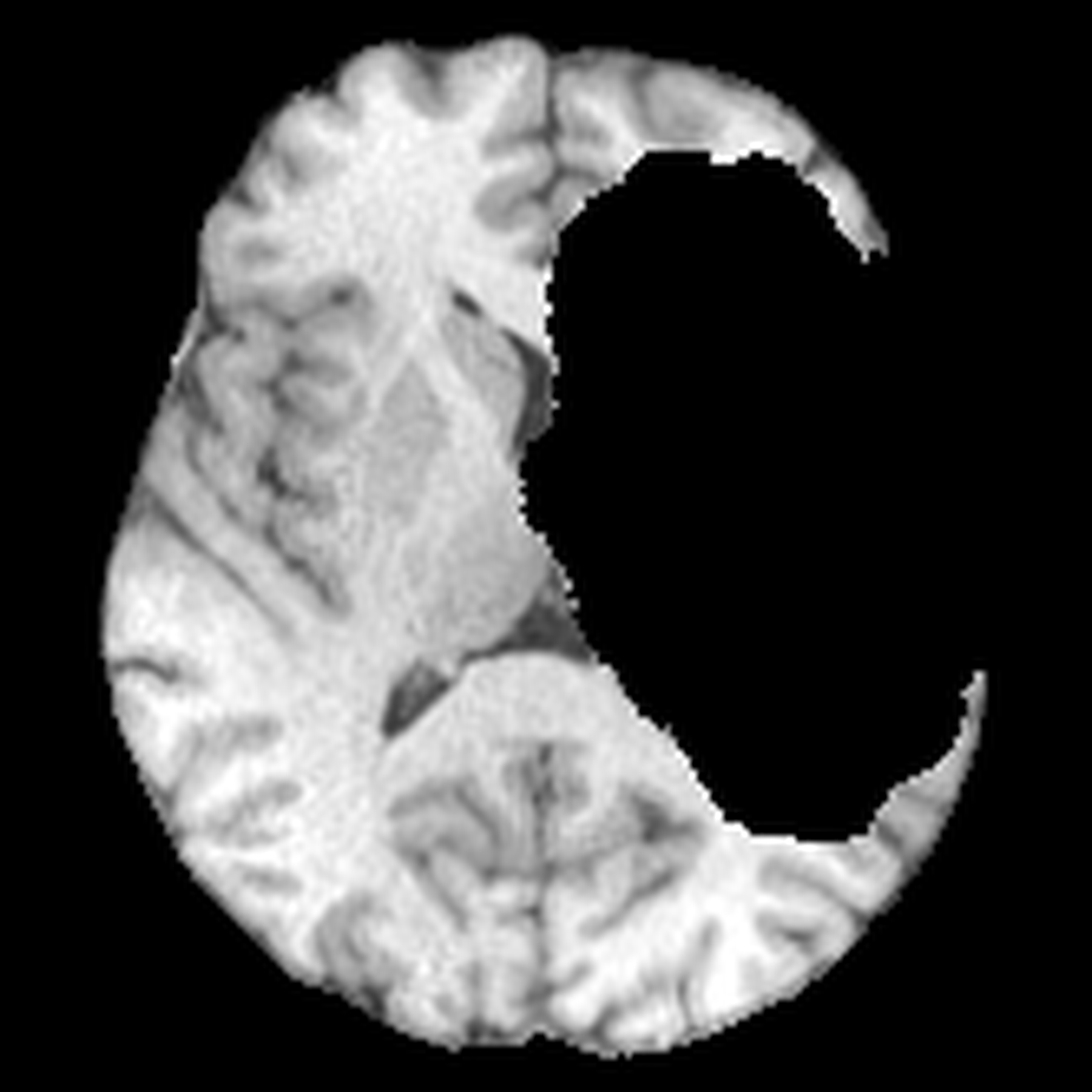}
\end{subfigure}
\hfill
\begin{subfigure}[t]{0.158\linewidth}
\centering
\includegraphics[width=\linewidth]{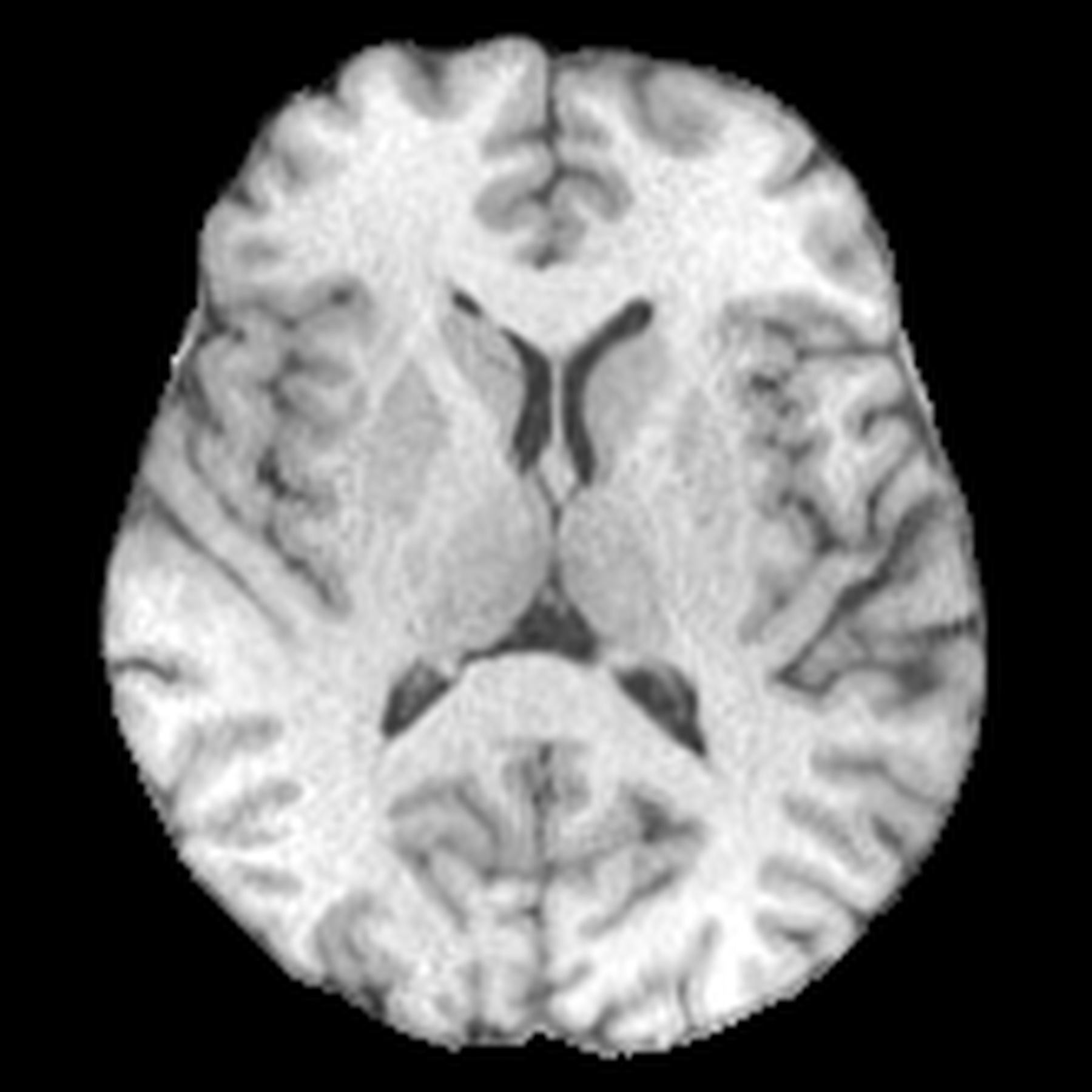}
\end{subfigure}
\hfill
\begin{subfigure}[t]{0.158\linewidth}
\centering
\includegraphics[width=\linewidth]{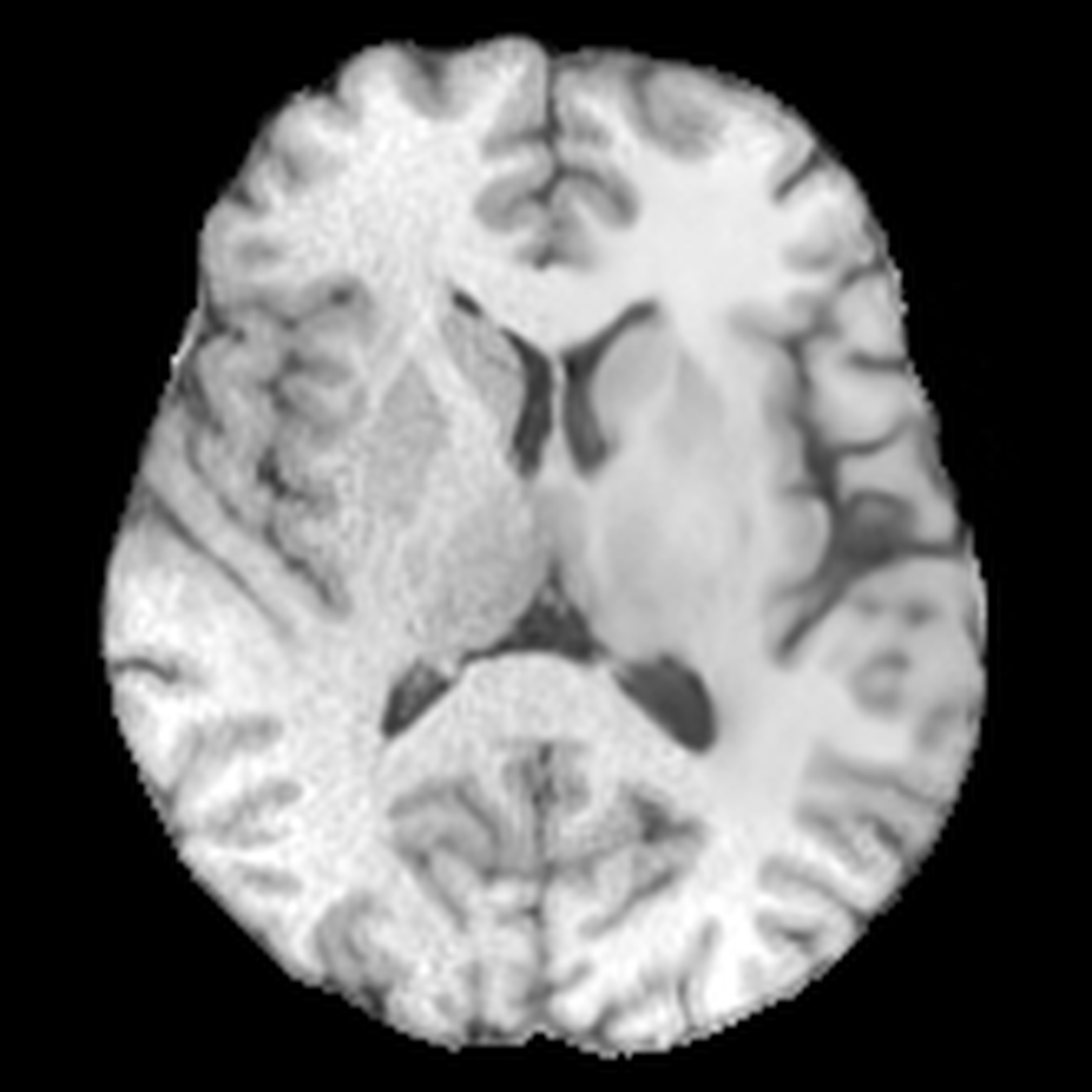}
\end{subfigure}
\hfill
\begin{subfigure}[t]{0.158\linewidth}
\centering
\includegraphics[width=\linewidth]{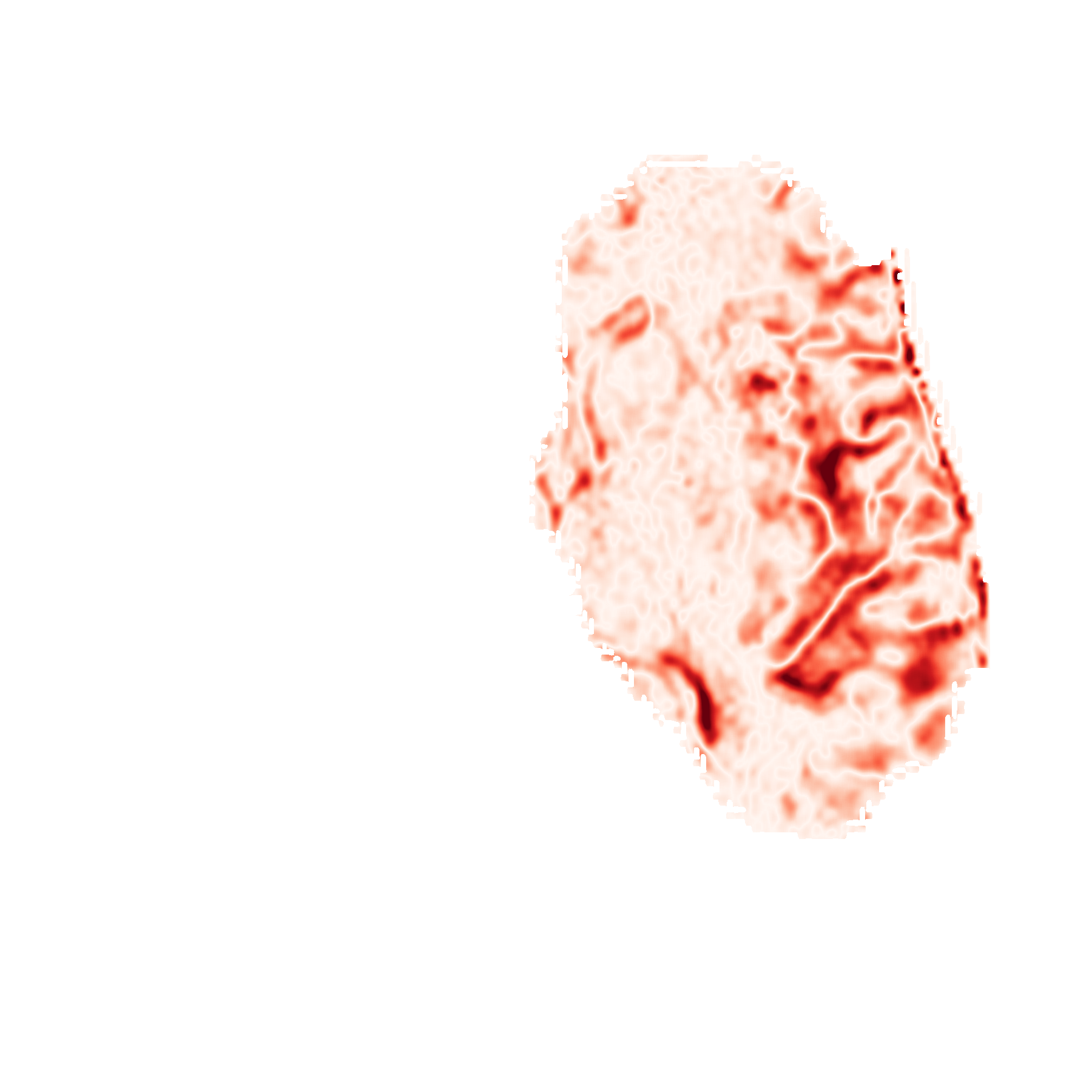}
\end{subfigure}
\hfill
\begin{subfigure}[t]{0.158\linewidth}
\centering
\includegraphics[width=\linewidth]{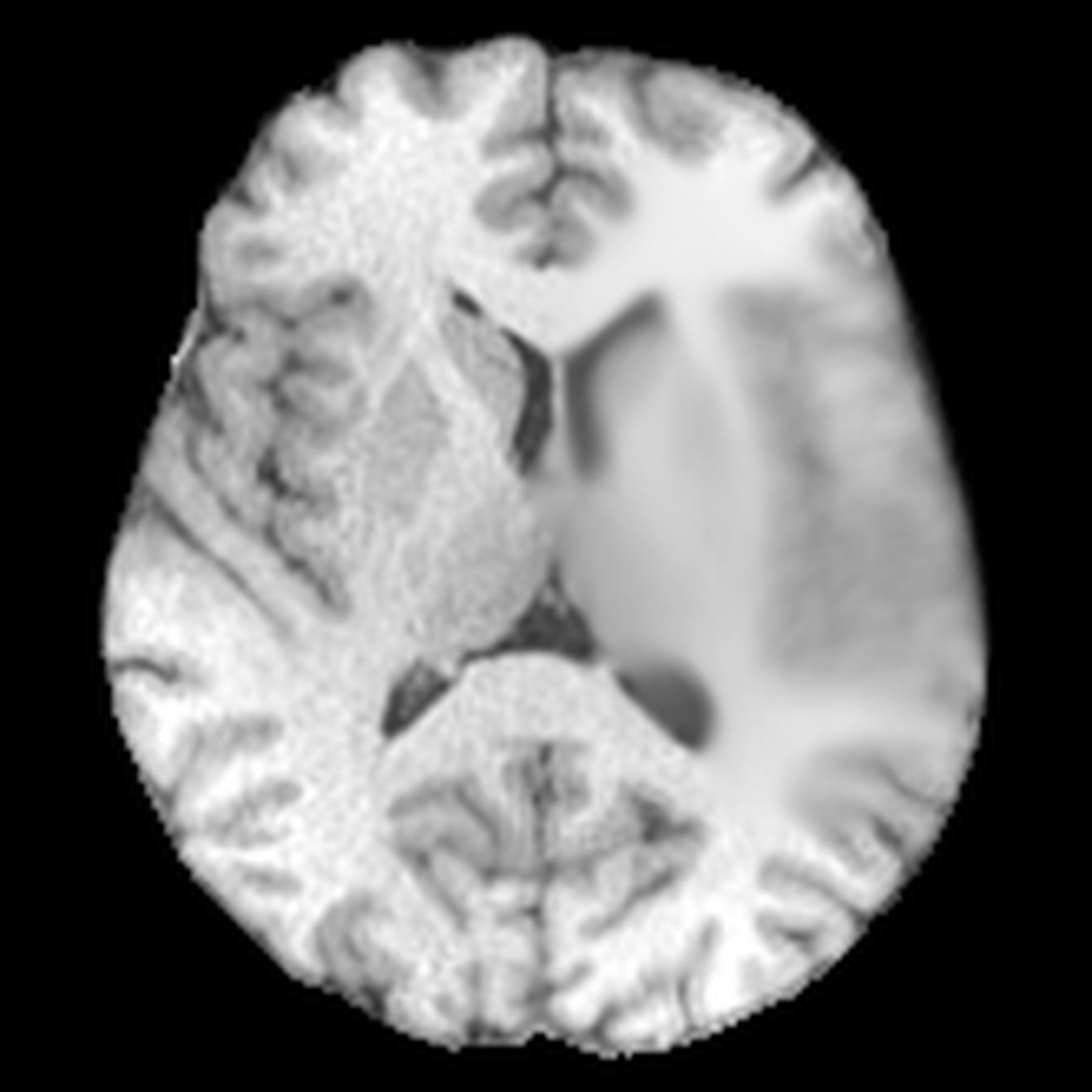}
\end{subfigure}
\hfill
\begin{subfigure}[t]{0.158\linewidth}
\centering
\includegraphics[width=\linewidth]{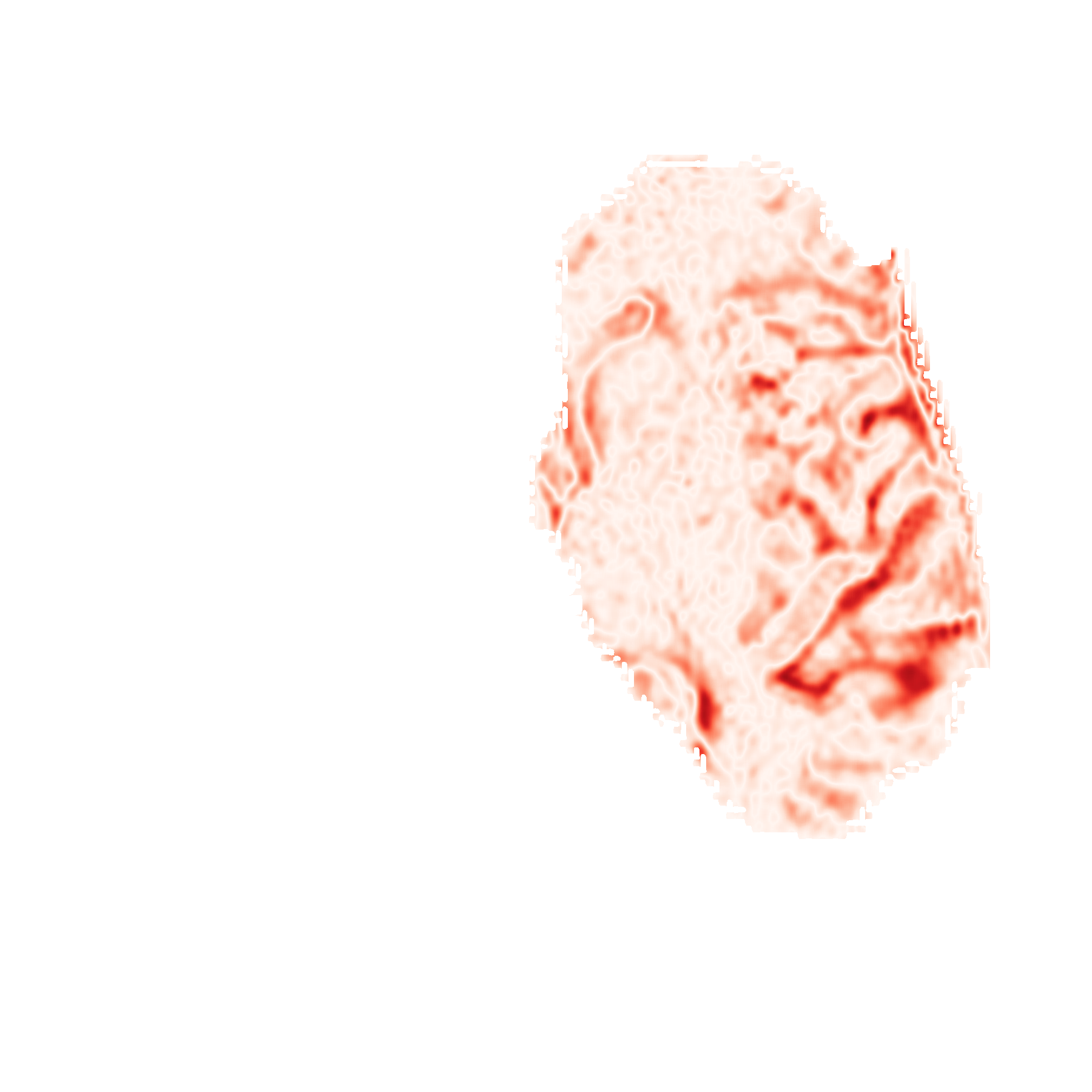}
\end{subfigure}

\par
\vspace{0.6em}

\begin{subfigure}[t]{0.158\linewidth}
\centering
\includegraphics[width=\linewidth]{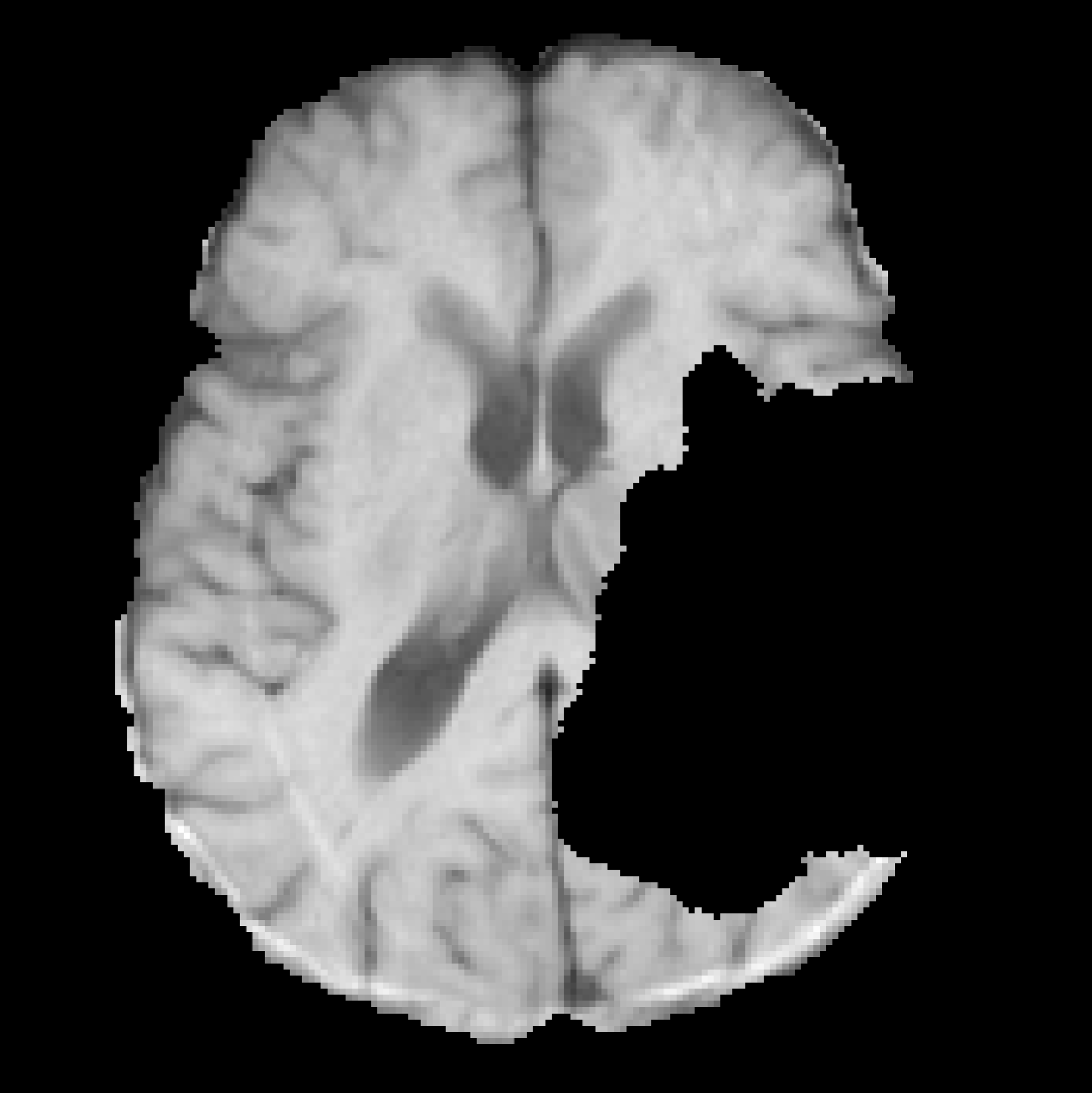}
\caption{\mbox{Voided input}}
\end{subfigure}
\hfill
\begin{subfigure}[t]{0.158\linewidth}
\centering
\includegraphics[width=\linewidth]{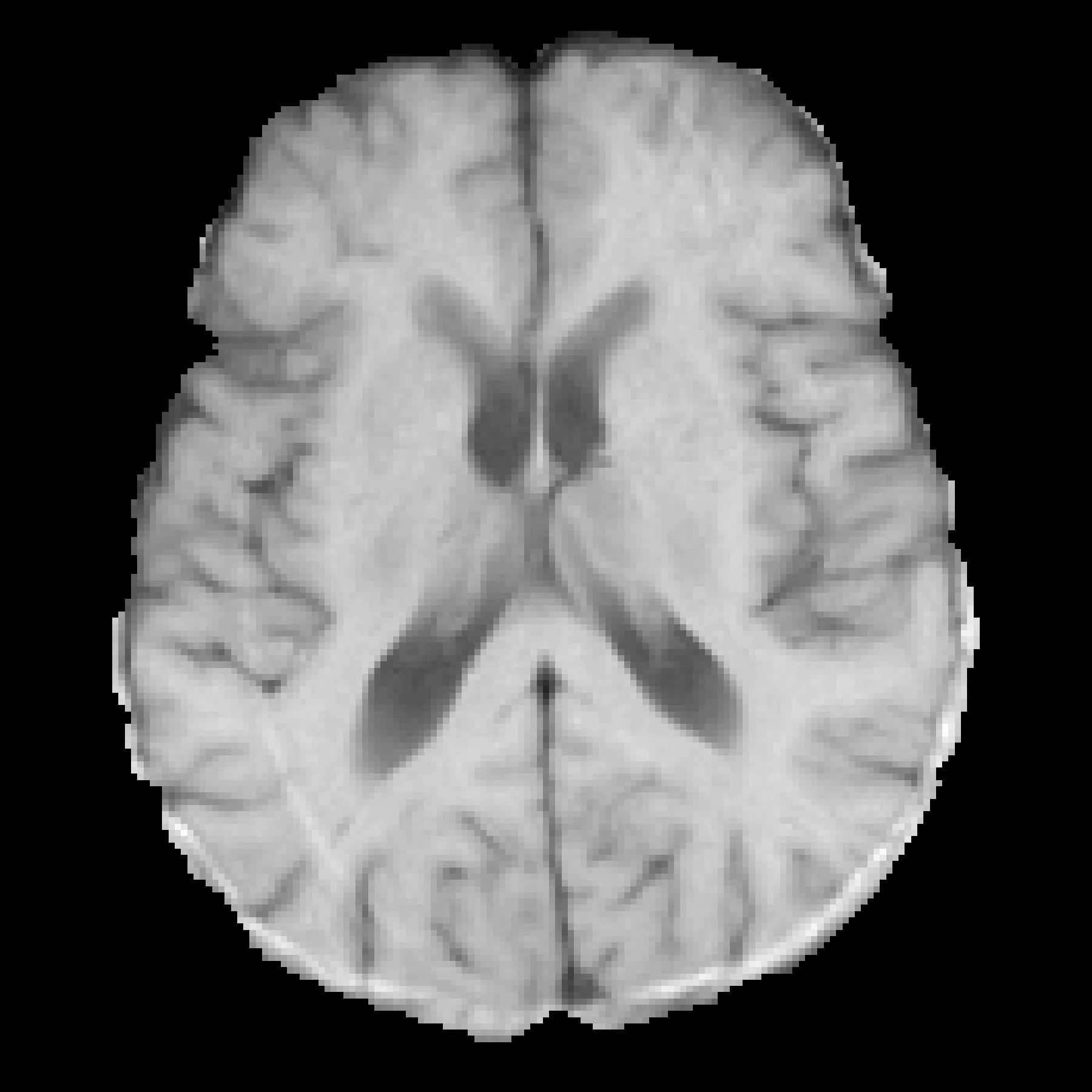}
\caption{\mbox{Ground truth}}
\end{subfigure}
\hfill
\begin{subfigure}[t]{0.158\linewidth}
\centering
\includegraphics[width=\linewidth]{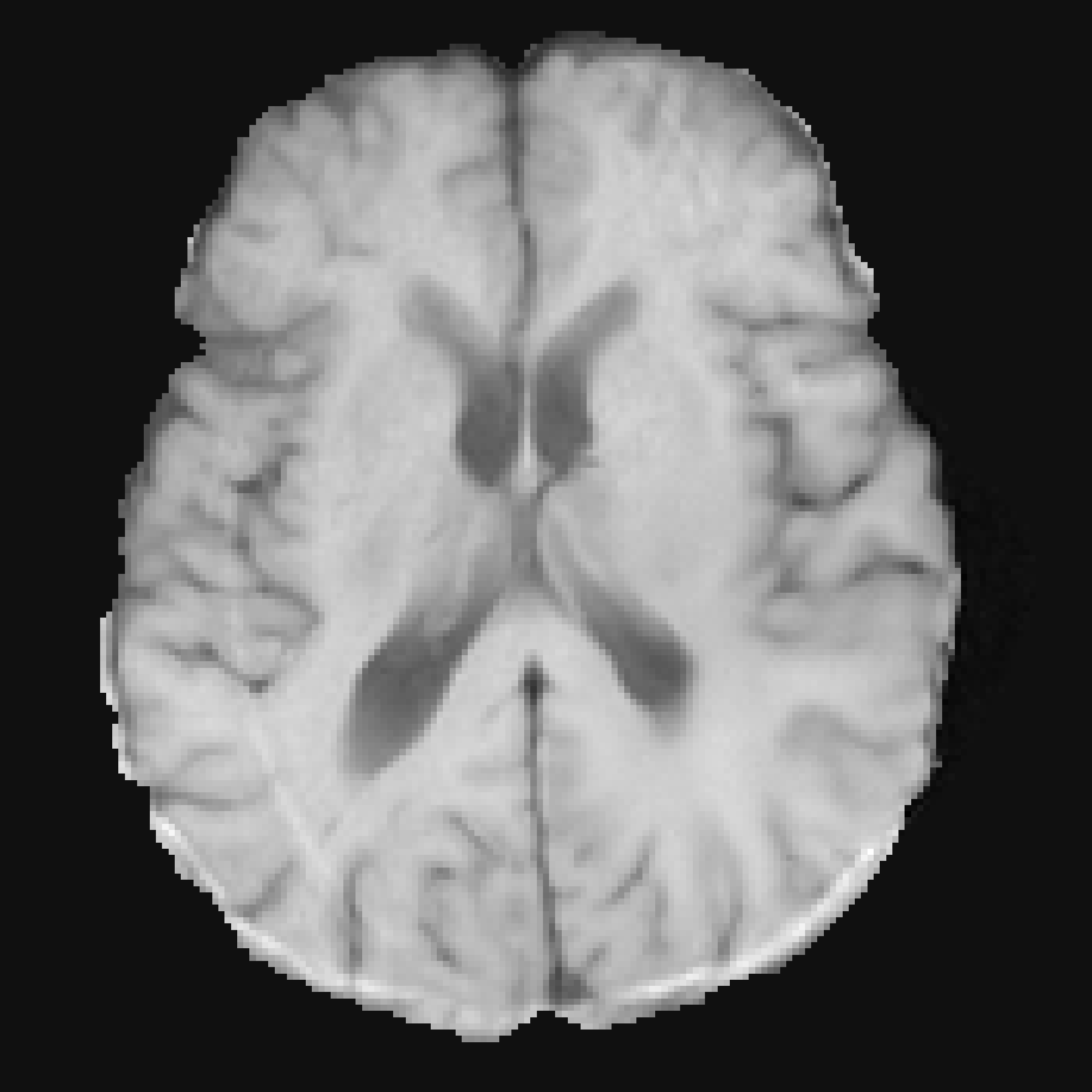}
\caption{\mbox{Single sample}}
\end{subfigure}
\hfill
\begin{subfigure}[t]{0.158\linewidth}
\centering
\includegraphics[width=\linewidth]{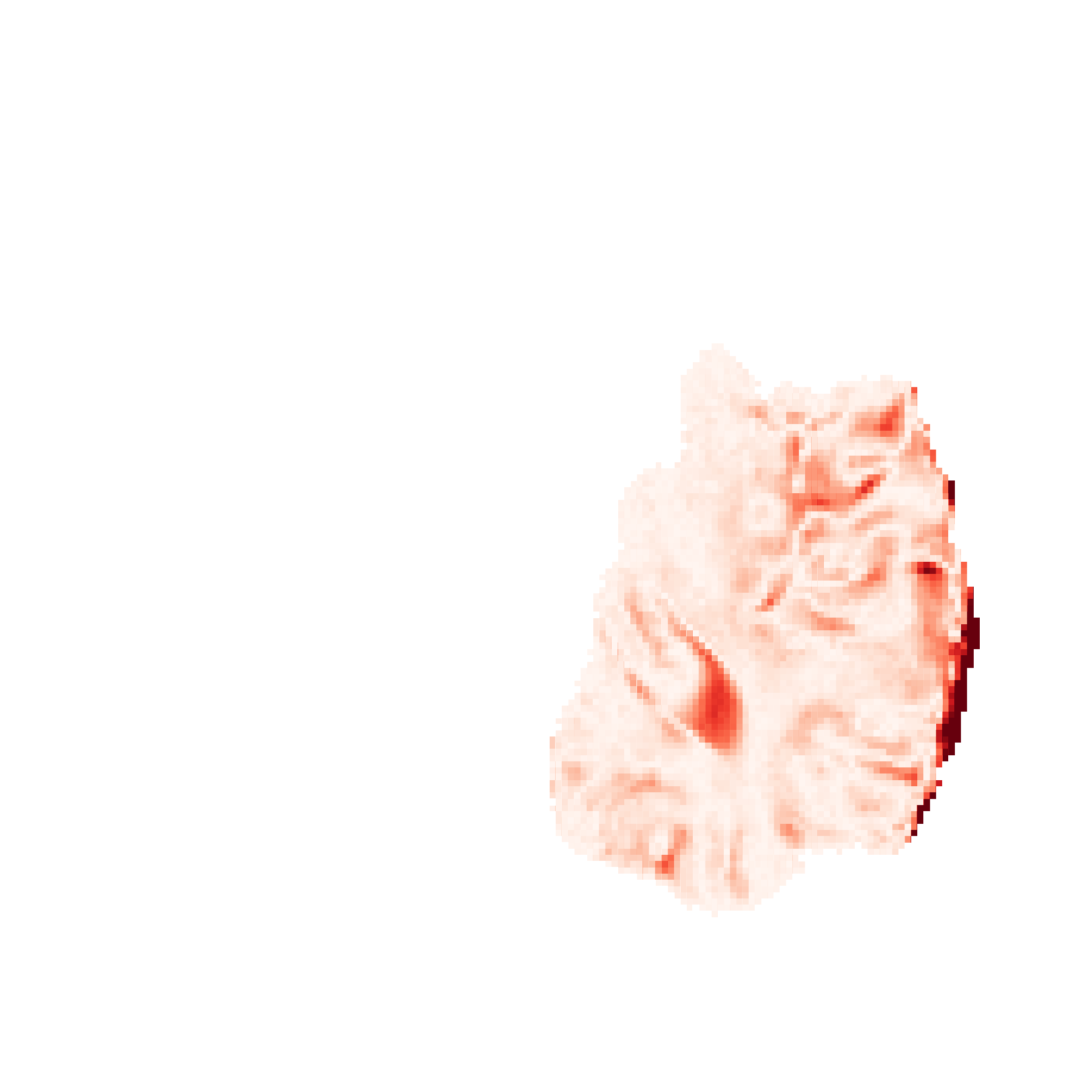}
\caption{\mbox{Single error}}
\end{subfigure}
\hfill
\begin{subfigure}[t]{0.158\linewidth}
\centering
\includegraphics[width=\linewidth]{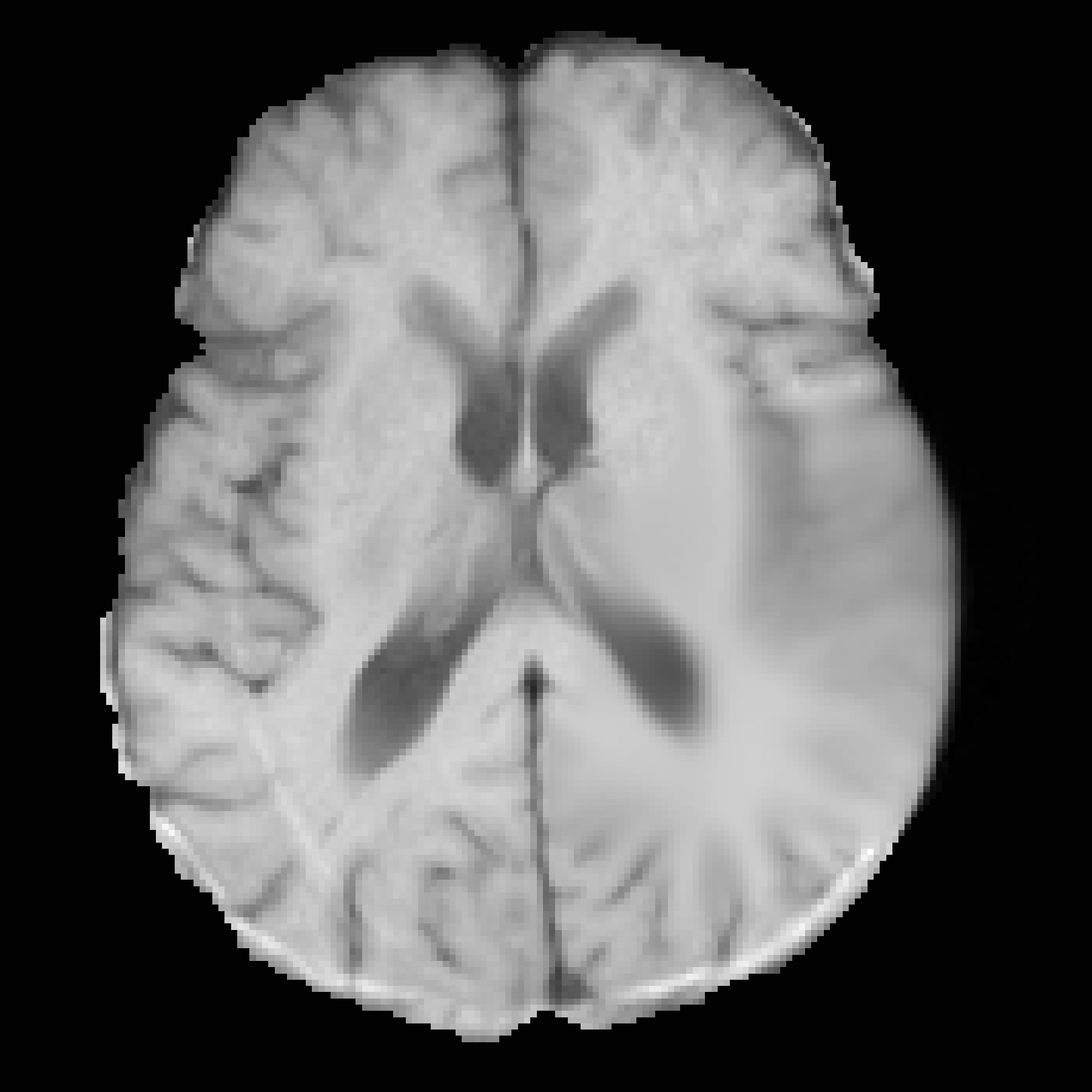}
\caption{\mbox{Mean}}
\end{subfigure}
\hfill
\begin{subfigure}[t]{0.158\linewidth}
\centering
\includegraphics[width=\linewidth]{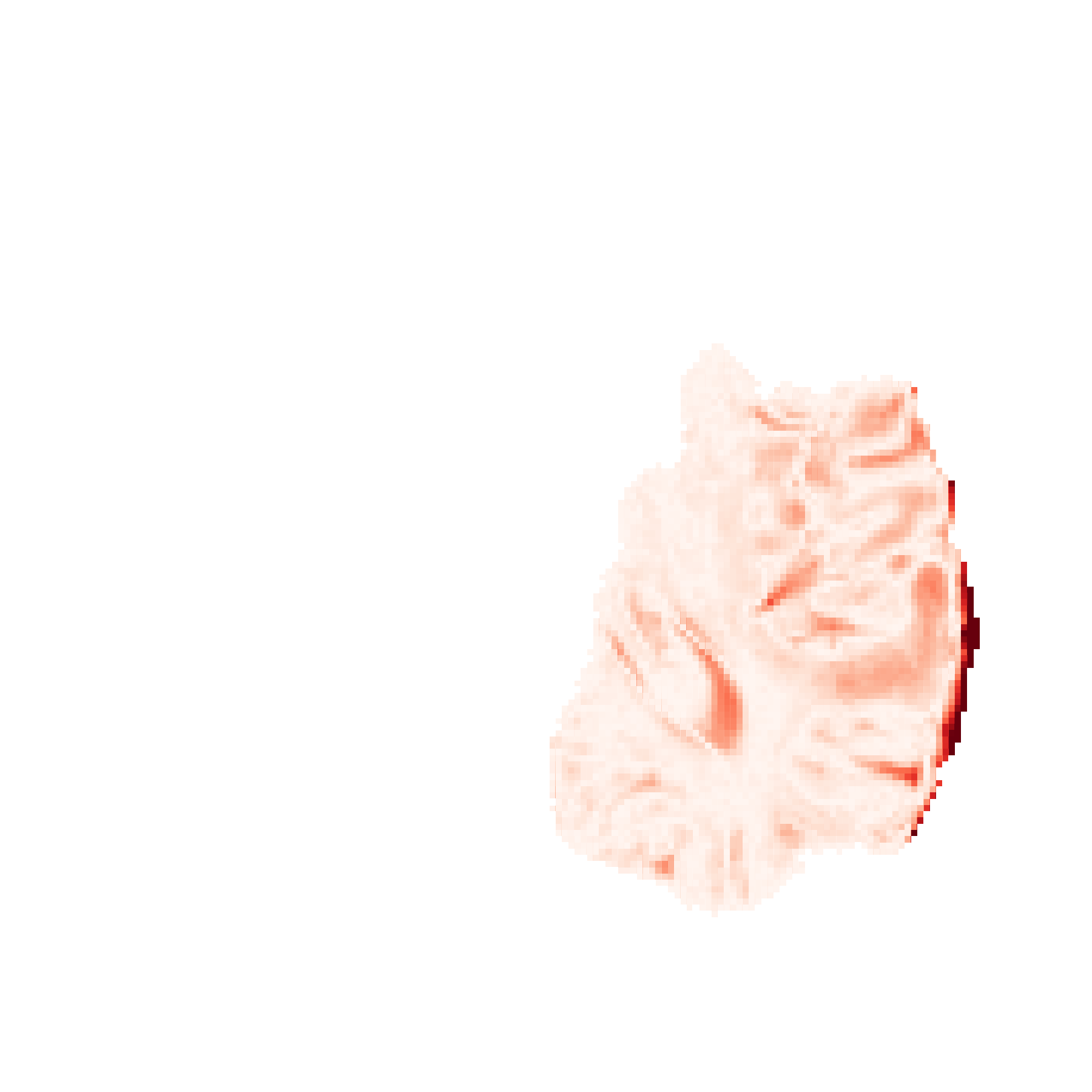}
\caption{\mbox{Mean error}}
\end{subfigure}

\end{minipage}
\hfill
\begin{minipage}[t]{0.11\textwidth}
\vspace{0pt}
\centering
\includegraphics[
    width=\linewidth
]{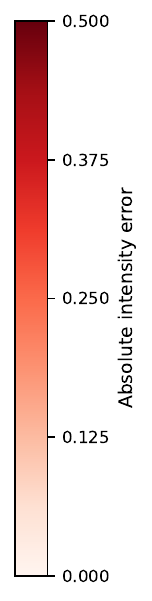}
\end{minipage}

\caption{
Qualitative comparison between single-sample inference and averaging
$K=30$ samples across three representative cases. Absolute-error maps are
restricted to the synthesized region and shown using a common color scale,
with white indicating zero error and increasingly saturated red indicating
larger discrepancies. Averaging generally improves distortion-based fidelity,
although it may smooth fine anatomical structures relative to individual
generated samples.
}
\label{fig:qualitative_ablation}
\end{figure*}

Figure~\ref{fig:qualitative_ablation} compares standard single-sample inference with \(K\)-sample averaging, together with the voided input and ground truth.

\subsection{Main challenge results}

\begin{table}[h]
\centering
\caption{Official BraTS validation performance of RARF using the released checkpoint and \(K=50\) sample averaging.}
\label{tab:main_results}
\begin{tabular}{ccc}
\hline
\textbf{MSE \(\downarrow\)}
& \textbf{PSNR \(\uparrow\)}
& \textbf{SSIM \(\uparrow\)} \\
\hline
0.006 & 24.008 & 0.832 \\
\hline

\end{tabular}
\end{table}

Table~\ref{tab:main_results} summarizes the official validation performance of our challenge submission.
This submission uses a separately trained checkpoint and averages  \(K=50\) samples per case to prioritize the distortion-based evaluation metrics. For reproducibility, we publicly release the corresponding model weights together with the source code. 

\section{Discussion}

RARF is conceived as a general region-aware inpainting framework rather than a method tailored exclusively to the BraTS benchmark. Its flexible formulation supports different training and inference configurations and can be applied to arbitrary inpainting masks. We believe this broader perspective is important, as medical imaging can both benefit from and contribute to advances in general computer vision.

Medical imaging also provides a particularly demanding benchmark domain because of its high dimensionality, complex spatial structure, and strict anatomical constraints. In inpainting, the masked region corresponds to a specific underlying anatomy rather than one of many equally valid completions. This makes reference-based distortion metrics more informative than in less constrained natural-image settings.

Nevertheless, our ablation shows that lower distortion does not necessarily imply better perceptual quality or anatomical plausibility. Averaging multiple samples improved MSE and PSNR but produced smoother, blurrier reconstructions that could still rank above sharper predictions. This motivates complementing voxel-wise metrics with perceptual, anatomy-aware, and downstream-task evaluation.

\begin{credits}
\subsubsection{\ackname} This study has been funded by the MAGERIT-CM project (TEC2024/COM-44), funded by Comunidad de Madrid.
 
\subsubsection{\discintname}
Authors declare no conflict of interests relevant to this research.
\end{credits}

%
%
%
%

\bibliographystyle{splncs04}
\bibliography{references}

@misc{kofler2024braintumorsegmentationbrats,
      title={The Brain Tumor Segmentation (BraTS) Challenge: Local Synthesis of Healthy Brain Tissue via Inpainting}, 
      author={Kofler, Florian and Meissen, Felix and Steinbauer, Felix and others},
      year={2024},
      eprint={2305.08992},
      archivePrefix={arXiv},
      primaryClass={eess.IV},
      url={https://arxiv.org/abs/2305.08992}, 
}

@misc{baid2021rsnaasnrmiccaibrats2021benchmark,
      title={The RSNA-ASNR-MICCAI BraTS 2021 Benchmark on Brain Tumor Segmentation and Radiogenomic Classification}, 
      author={Baid, Ujjwal and Ghodasara, Satyam and Mohan, Suyash and others},
      year={2021},
      eprint={2107.02314},
      archivePrefix={arXiv},
      primaryClass={cs.CV},
      url={https://arxiv.org/abs/2107.02314}, 
}

@article{pollak2025fastsurferlit,
  title   = {{FastSurfer-LIT}: Lesion inpainting tool for whole-brain {MRI} segmentation with tumors, cavities, and abnormalities},
  author  = {Pollak, Christoph and K{\"u}gler, David and Bauer, Tobias and R{\"u}ber, Thomas and Reuter, Martin},
  journal = {Imaging Neuroscience},
  volume  = {3},
  pages   = {imag\_a\_00446},
  year    = {2025},
  month   = jan,
  doi     = {10.1162/imag_a_00446},
  pmid    = {40109899},
  pmcid   = {PMC11917724}
}

@inproceedings{bertalmio2000imageinpainting,
author = {Bertalmío, Marcelo and Sapiro, Guillermo and Caselles, Vicent and Ballester, C.},
year = {2000},
month = {01},
pages = {417-424},
title = {Image inpainting},
journal = {Proceedings of the ACM SIGGRAPH Conference on Computer Graphics}
}

@article{telea2004fastmarching,
author = {Telea, Alexandru},
year = {2004},
month = {01},
pages = {},
title = {An Image Inpainting Technique Based on the Fast Marching Method},
volume = {9},
journal = {Journal of Graphics Tools},
doi = {10.1080/10867651.2004.10487596}
}

@ARTICLE{criminisi2004exemplar,
  author={Criminisi, A. and Perez, P. and Toyama, K.},
  journal={IEEE Transactions on Image Processing}, 
  title={Region filling and object removal by exemplar-based image inpainting}, 
  year={2004},
  volume={13},
  number={9},
  pages={1200-1212},
  doi={10.1109/TIP.2004.833105}}

@InProceedings{10.1007/978-3-030-01252-6_6,
author="Liu, Guilin
and Reda, Fitsum A.
and Shih, Kevin J.
and Wang, Ting-Chun
and Tao, Andrew
and Catanzaro, Bryan",
title="Image Inpainting for Irregular Holes Using Partial Convolutions",
booktitle="Computer Vision -- ECCV 2018",
year="2018",
publisher="Springer International Publishing",
address="Cham",
pages="89--105",
isbn="978-3-030-01252-6"
}

@INPROCEEDINGS{9880056,
  author={Lugmayr, Andreas and Danelljan, Martin and Romero, Andres and Yu, Fisher and Timofte, Radu and Van Gool, Luc},
  booktitle={2022 IEEE/CVF Conference on Computer Vision and Pattern Recognition (CVPR)}, 
  title={RePaint: Inpainting using Denoising Diffusion Probabilistic Models}, 
  year={2022},
  volume={},
  number={},
  pages={11451-11461},
  doi={10.1109/CVPR52688.2022.01117}}

@inproceedings{armanious2020ipa,
  title={ipA-MedGAN: Inpainting of arbitrary regions in medical imaging},
  author={Armanious, Karim and Kumar, Vijeth and Abdulatif, Sherif and Hepp, Tobias and Gatidis, Sergios and Yang, Bin},
  booktitle={2020 IEEE international conference on image processing (ICIP)},
  pages={3005--3009},
  year={2020},
  organization={IEEE}
}

@InProceedings{10.1007/978-3-030-59520-3_5,
author="Manj{\'o}n, Jos{\'e} V.
and Romero, Jos{\'e} E.
and Vivo-Hernando, Roberto
and Rubio, Gregorio
and Aparici, Fernando
and de la Iglesia-Vaya, Maria
and Tourdias, Thomas
and Coup{\'e}, Pierrick",
title="Blind MRI Brain Lesion Inpainting Using Deep Learning",
booktitle="Simulation and Synthesis in Medical Imaging",
year="2020",
publisher="Springer International Publishing",
address="Cham",
pages="41--49",
isbn="978-3-030-59520-3"
}

@inproceedings{durrer2024denoising,
  title={Denoising diffusion models for 3d healthy brain tissue inpainting},
  author={Durrer, Alicia and Wolleb, Julia and Bieder, Florentin and Friedrich, Paul and others},
  booktitle={MICCAI Workshop on Deep Generative Models},
  pages={87--97},
  year={2024},
  organization={Springer}
}

@InProceedings{zhang2025robust,
author="Zhang, Juexin
and Weng, Ying
and Chen, Ke",
title="Robust 3D Brain MRI Inpainting with Random Masking Augmentation",
booktitle="Segmentation, Classification, and Synthesis for Brain Tumors and Traumatic Brain Injuries",
year="2026",
publisher="Springer Nature Switzerland",
address="Cham",
pages="102--109",
isbn="978-3-032-16370-7"
}

@InProceedings{zhang2024synthesis,
author="Zhang, Juexin
and Chen, Ke
and Weng, Ying",
title="Synthesis of Healthy Tissue Within Tumor Area via U-Net",
booktitle="Brain Tumor Segmentation, and Cross-Modality Domain Adaptation for Medical Image Segmentation",
year="2024",
publisher="Springer Nature Switzerland",
address="Cham",
pages="233--240",
isbn="978-3-031-76163-8"
}

@article{zhang2025unet,
  title={U-Net Based Healthy 3D Brain Tissue Inpainting},
  author={Juexin Zhang and Ying Weng and Ke Chen},
  journal={ArXiv},
  year={2025},
  volume={abs/2507.18126},
  url={https://api.semanticscholar.org/CorpusID:280017988}
}

@article{Liu2022FlowSA,
  title={Flow Straight and Fast: Learning to Generate and Transfer Data with Rectified Flow},
  author={Xingchao Liu and Chengyue Gong and Qiang Liu},
  journal={ArXiv},
  year={2022},
  volume={abs/2209.03003},
  url={https://api.semanticscholar.org/CorpusID:252111177}
}

@INPROCEEDINGS{kim2025rad,
  author={Kim, Sora and Suh, Sungho and Lee, Minsik},
  booktitle={2025 IEEE/CVF Conference on Computer Vision and Pattern Recognition (CVPR)}, 
  title={RAD: Region-Aware Diffusion Models for Image Inpainting}, 
  year={2025},
  volume={},
  number={},
  pages={2439-2448},
  doi={10.1109/CVPR52734.2025.00233}}

@inproceedings{DBLP:conf/nips/HuMSKMD25,
  author       = {Yuyang Hu and
                  Kangfu Mei and
                  Mojtaba Sahraee{-}Ardakan and
                  Ulugbek Kamilov and
                  Peyman Milanfar and
                  Mauricio Delbracio},
  title        = {Kernel Density Steering: Inference-Time Scaling via Mode Seeking for
                  Image Restoration},
  booktitle    = {Advances in Neural Information Processing Systems 38: Annual Conference
                  on Neural Information Processing Systems 2025, NeurIPS 2025, San Diego,
                  CA, USA, December 2-7, 2025 / Mexico City, Mexico, November 30 - December
                  5, 2025},
  year         = {2025},
  bibsource    = {dblp computer science bibliography, https://dblp.org}
}

@inproceedings{natsumi-etal-2025-agreement,
    title = "Agreement-Constrained Probabilistic Minimum {B}ayes Risk Decoding",
    author = "Natsumi, Koki  and
      Deguchi, Hiroyuki  and
      Sakai, Yusuke  and
      Kamigaito, Hidetaka  and
      Watanabe, Taro",
    booktitle = "Proceedings of the 14th International Joint Conference on Natural Language Processing and the 4th Conference of the Asia-Pacific Chapter of the Association for Computational Linguistics",
    month = dec,
    year = "2025",
    address = "Mumbai, India",
    publisher = "The Asian Federation of Natural Language Processing and The Association for Computational Linguistics",
    doi = "10.18653/v1/2025.ijcnlp-short.39",
    pages = "484--493",
    ISBN = "979-8-89176-299-2",
}

@inproceedings{NEURIPS2021_d77e6859,
 author = {Freirich, Dror and Michaeli, Tomer and Meir, Ron},
 booktitle = {Advances in Neural Information Processing Systems},
 pages = {25661--25672},
 publisher = {Curran Associates, Inc.},
 title = {A Theory of the Distortion-Perception Tradeoff in Wasserstein Space},
 volume = {34},
 year = {2021}
}

@unknown{torrado,
author = {Torrado-Carvajal, Angel and Albrecht, Daniel and Lee, Jeungchan and Andronesi, Ovidiu and Ratai, Eva-Maria and Napadow, Vitaly and Loggia, Marco},
year = {2020},
month = {02},
pages = {},
title = {Inpainting as a Technique for Estimation of Missing Voxels in Chemical Shift Imaging},
doi = {10.1101/2020.02.17.952325}
}

\end{document}